\documentclass[letterpaper, 10 pt, conference]{ieeeconf}  

\IEEEoverridecommandlockouts                              

\usepackage{amsmath,amsfonts}
\usepackage{graphicx}
\usepackage{cite}

\usepackage{graphics} 
\usepackage{epsfig}
\usepackage{times} 
\usepackage{amssymb}  
\usepackage{booktabs}
\usepackage{multirow}
\usepackage{soul}
\usepackage[table,xcdraw]{xcolor}
\usepackage{flushend}
\usepackage{subfigure}
\usepackage{siunitx}
\usepackage{ulem}

\soulregister\cite7 
\soulregister\citep7 
\soulregister\citet7 
\soulregister\ref7 
\soulregister\pageref7 

\title{\LARGE \bf
Learning Natural and Robust Hexapod Locomotion over Complex Terrains via Motion Priors based on Deep Reinforcement Learning
}

\author{Xin Liu, Jinze Wu, Yinghui Li, Chenkun Qi$^{*}$, Yufei Xue, Feng Gao 
\thanks{This work was supported by the National Key Research and Development Plan (2021YFF0307900). All authors are with School of Mechanical Engineering, Shanghai Jiao Tong University, Shanghai, China. Email: {\tt\small chenkqi@sjtu.edu.cn}}
\thanks{$^{*}$ Corresponding Author}
}

\begin{document}

\maketitle
\thispagestyle{empty}
\pagestyle{empty}

\begin{abstract}

Multi-legged robots offer enhanced stability to navigate complex terrains with their multiple legs interacting with the environment. However, how to effectively coordinate the multiple legs in a larger action exploration space to generate natural and robust movements is a key issue. In this paper, we introduce a motion prior-based approach, successfully applying deep reinforcement learning algorithms to a real hexapod robot. We generate a dataset of optimized motion priors, and train an adversarial discriminator based on the priors to guide the hexapod robot to learn natural gaits. The learned policy is then successfully transferred to a real hexapod robot, and demonstrate natural gait patterns and remarkable robustness without visual information in complex terrains. This is the first time that a reinforcement learning controller has been used to achieve complex terrain walking on a real hexapod robot.

\end{abstract}


\section{INTRODUCTION}

Hexapod robots, like their natural counterparts, are known for superior terrain adaptability and stability, with their quasi-static gait requiring minimal muscle output \cite{Zhong7876812}. As a result, they have gained significant interest and application. However, developing a controller for natural gait and robust motion on complex terrains remains a challenge.

Previous research on hexapod robots often relies on static gaits, like crawling, which limits their ability to navigate challenging terrains quickly and reliably. Locomotion controllers for bipedal and quadrupedal robots focus on two main types: model-based, which rely on simplified environmental and robot dynamics, and model-free, which use data-driven approaches without explicit modeling.

Model-free deep reinforcement learning (DRL) algorithms have proven more robust in complex environments compared to model-based methods, leading to increased use of DRL in legged locomotion control \cite{schilling2020decentralized, hwangbo2019learning, lee2020learning, wjz, ji2022concurrent}. Despite this, no DRL algorithm has yet been applied effectively to real hexapod robots for natural and robust locomotion in complex terrains. The challenge lies in the increased complexity due to more legs, making it harder for robots to generate natural gaits. This work presents a DRL-based method for hexapod motion control, incorporating motion priors, to enable robust, natural locomotion in challenging environments.

\begin{figure}[tbp]
\includegraphics[width=\linewidth]{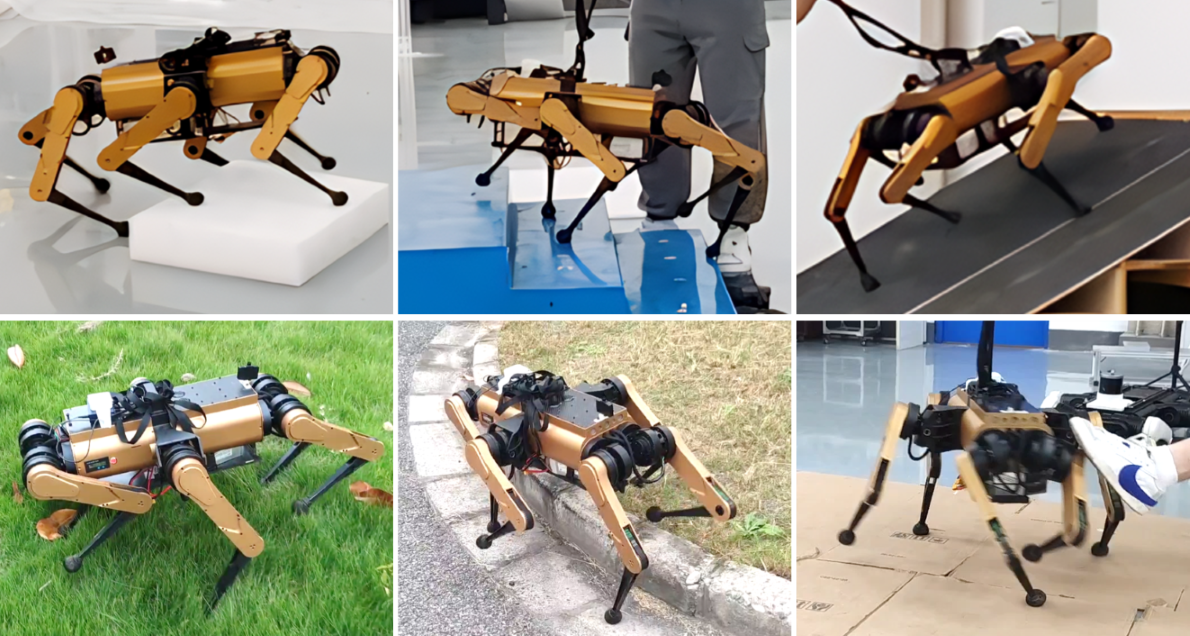}
\centering
\caption{The hexapod robot showcases its ability to achieve natural and robust locomotion across diverse terrains.}
\label{fig:locomotion}
\vspace{-0.5em}
\end{figure}

The main contributions are listed as follows:
\begin{enumerate}
\item We produce motion data for the hexapod robot on flat terrain using trajectory optimization (TO). Subsequently, we trained a motion discriminator to assist the hexapod robot in achieving a natural and robust locomotion in challenging terrains.
\item We propose an asymmetric DRL framework based on adversarial discriminator for training a motion controller and deploy it on a real hexapod robot to achieve blind locomotion in challenging terrains.
\end{enumerate}

\section{RELATED WORK}

\subsection{Locomotion Control Algorithms for Legged Robots}

Researchers have studied legged robot motion control to enable adaptation to complex terrains. Model-based methods, such as model predictive control (MPC) \cite{meduri2023biconmp} and whole body control (WBC) \cite{mit2019wbc}, require simplification and modeling of robot dynamics and the environment. However, these approaches struggle with unstructured or unknown terrains, which can lead to optimization failures.

An alternative approach incorporates biological concepts like central pattern generators (CPG) into control methods to reduce task complexity \cite{schilling2020decentralized, lele2020learning}. However, adjusting CPG parameters online in changing environments is challenging, often compromising stability, especially in dynamic or unknown conditions.

Recently, data-driven algorithms, particularly reinforcement learning (RL), have gained popularity for controlling bipedal and quadrupedal robots \cite{hwangbo2019learning, lee2020learning, miki2022learning, Liu2024Skill}. These methods rely on proprioceptive sensors, like joint encoders and the IMU, offering a more robust solution for unstructured environments. However, robots with more legs face increased difficulty in learning natural and stable gaits due to a larger exploration space, making convergence harder and reward function design more complex.

\subsection{Reinforcement Learning for Locomotion}

RL controllers have proven effective for legged robots, especially quadrupeds, enhancing their motion capabilities and adaptability to complex terrains \cite{hwangbo2019learning, lee2020learning, wjz, ji2022concurrent}. \cite{hwangbo2019learning} used an actuator network to model actuator dynamics and ensure smooth transition from simulation to reality. Building on this, \cite{lee2020learning} improved ANYmal's robustness by training it on various terrains. Some RL controllers for bipedal robots, like Cassie, adjust reference motions from a pre-defined model-based controller \cite{xie2018feedback, xie2020learning}, speeding up training but limiting motion flexibility and exploration.

Hexapod robots, with more points of contact with the ground, offer better stability and interaction with the terrain, allowing greater perception of terrain complexity. However, most RL research on hexapods focuses on crawling gaits \cite{schilling2020decentralized, lele2020learning}, limiting agility and speed in complex environments. \cite{azayev2020blind} proposed a scalable two-level framework for blind hexapod locomotion in complex environments using RL, training expert policies on discrete terrain distributions. However, this method has only been tested in simulations. Currently, no RL framework exists for real hexapod robots to learn natural, robust gaits for challenging terrains using only proprioception.

\subsection{Motion Imitation Learning}

Designing complex reward functions is laborious, especially for hexapod robots exploring higher-dimensional spaces. Achieving a natural, robust gait via meticulously crafted reward functions is challenging. Imitation learning offers an alternative: by imitating real animal motion or manually crafted animation data, learning can converge faster and achieve higher-quality performance \cite{peng2018deepmimic, peng2020learning}. However, while this approach effectively replicates individual motion clips, it struggles to handle multiple reference motions with a single phase variable.

Adversarial Motion Priors (AMP) \cite{peng2021amp} address this issue using a GAIL framework \cite{ho2016generative} that builds an adversarial discriminator. The discriminator discerns whether state transition pairs \(\left(s_t,s_{t+1}\right)\) come from prior data or the learned policy, guiding the agent toward the motion characteristics of the prior data.
This approach allows simulated agents to perform complex tasks while adopting motion styles from large, unstructured motion datasets \cite{wjz, escontrela2022adversarial, vollenweider2022advanced}.


In this work, we employ a more general motion imitation approach based on adversarial imitation learning and construct an asymmetric reinforcement learning network. This enables it to be trained using privileged information in simulation, relying solely on proprioceptive sensors for zero-shot generalization to the real hexapod robot without the need for fine-tuning. This allows our hexapod robot to exhibit similar behavior to a raw motion dataset on flat terrain without motion clips and to adapt to challenging terrains.

\section{LEARNING FROM MOTION PRIORS}\label{sec:LEARNING_FROM_MOTION_PRIORS}

\begin{figure*}[htbp]
\includegraphics[width=0.85\linewidth]{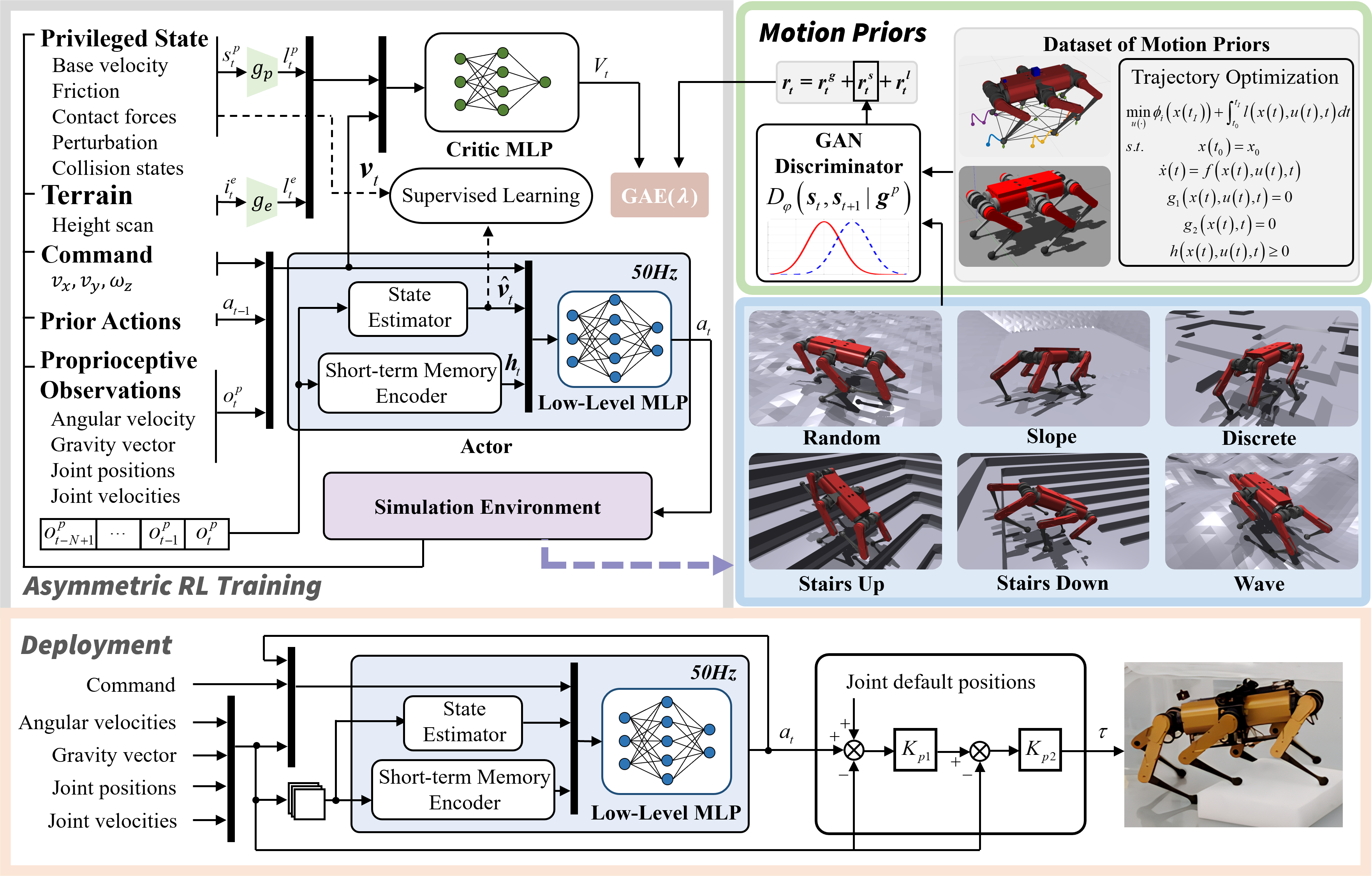}
\centering
\caption{The asymmetric Actor-Critic reinforcement learning framework. We formulate three types of rewards to facilitate tripod gait styles. The style-specific reward is given by the discriminator of adversarial motion priors. During deployment, the desired joint position calculated by summing the policy output with the default joint position is sent to the CSP controller to calculate the torque.}
\label{fig:method}
\vspace{-0.5em}
\end{figure*}

We consider a discrete-time dynamic model. At each time step $t$, the state is $\boldsymbol{x}_t$. An action $\boldsymbol{a}_t$ is taken according to the policy, leading to the next state $\boldsymbol{x}_{t+1}$ with probability \(P\left({{\boldsymbol{x}_{t + 1}}\mid{\boldsymbol{x}_t},{\boldsymbol{a}_t}}\right)\) and yielding a reward $r_t$. The goal of RL is to learn a policy parameterized by $\theta$, denoted \({\pi _\theta }\), that maximizes the discounted cumulative return: $J\left ( \theta  \right ) =\mathbb{E} _{\pi _{\theta } } \left (  {\textstyle \sum_{t=0}^{\infty }} \gamma ^{t}r_{t}   \right ) $.
Our controller does not use exteroreception, so the robot cannot obtain terrain data from cameras or radars. Consequently, the problem is modeled as a partially observable markov decision process (POMDP). We employ an asymmetric Actor-Critic framework \cite{asymmetricAC} to train the controller: the Critic has full access to the state (including terrain and privileged robot data), while the Actor can only access partial observations from proprioceptive sensors.

\textbf{Observation and Action Space:}  As shown in Fig. \ref{fig:method}, the Actor and Critic receive different inputs, reflecting their asymmetric roles. The Critic’s input includes comprehensive state observations for evaluating the Actor’s actions: proprioceptive data \(\boldsymbol{o}_t^p\in {\mathbb{R}^{42}}\), the previous action \(\boldsymbol{a}_{t-1}\in {\mathbb{R}^{18}}\), the target base velocity $\boldsymbol{v}_t^{\rm des}=\left(v_x,v_y,\omega_z\right)\in {\mathbb{R}^{3}}$, privileged state data \(\boldsymbol{s}_t^p\in {\mathbb{R}^{42}}\), and terrain elevation scanning points \(\boldsymbol{i}_t^e \in {\mathbb{R}^{187}}\). Proprioceptive data consists of the robot’s angular velocities \(\boldsymbol{\omega}_t\in {\mathbb{R}^{3}}\), gravity vector projection \(\boldsymbol{e}_g\in {\mathbb{R}^{3}}\), joint positions \(\boldsymbol{\theta}_t\in {\mathbb{R}^{18}}\), and joint velocities \(\boldsymbol{\dot{\theta}}_t\in {\mathbb{R}^{18}}\). Privileged state data includes base velocity \(\boldsymbol{v}_t\in {\mathbb{R}^{3}}\), base height \({h}_b\in {\mathbb{R}^{}}\), ground friction \({f}_n\in {\mathbb{R}^{}}\), foot contact forces \(\boldsymbol{f}_c\in {\mathbb{R}^{18}}\), external perturbation and its direction \(\boldsymbol{f}_p\in {\mathbb{R}^{6}}\), and collision states of the trunk, thighs, and calves \({\mathbb{I}}_c\in {\mathbb{R}^{13}}\), which are less directly measurable. Terrain information is collected from multiple surrounding points, indicating vertical displacement from the robot’s base.
By contrast, the Actor’s input is limited to proprioceptive data, the previous action, and the target base velocity. The policy action $\boldsymbol{a}_{t}$ is an 18-dimensional vector specifying a target joint position offset. This offset is added to the nominal joint position $\boldsymbol{q}_0$, which remains constant, to determine the desired motor position $\boldsymbol{q}_d$. The following low-level joint CSP control law then computes torques: $\boldsymbol{\tau}=\boldsymbol{K}_{p2}\left(\boldsymbol{K}_{p1}\left(\boldsymbol{q}_d-\boldsymbol{q}\right)-\boldsymbol{\dot{q}}\right)$.

\textbf{Reward Design:} Designing reward functions for hexapod robots can be challenging and requires expert tuning. When using rewards from quadrupeds, hexapods typically fail to develop the tripod gait. To address this, we design a reward with three components: a task tracking reward \(r^g_t\), a penalty \(r^l_t\), and a tripod-style reward \(r^s_t\). Their sum forms the total reward $ \label{eq:rewardDefinition}r_t = r^g_t + r^s_t + r^l_t$.
The task reward emphasizes accurate tracking of linear and angular velocities. The penalty promotes motion stability, smoothness, and safety. Specifically, penalties are applied to the body’s vertical velocity and roll/pitch angular velocities to maintain stability. Excessive joint torque and acceleration are penalized to reduce motor stress and conserve energy. The rate of action change is penalized for smooth motion. Joint torque and velocity exceeding thresholds are penalized to prevent hardware overload. Collisions and contact forces are penalized to encourage minimal collisions and prevent excessive body damage. The tripod-style reward, based on adversarial motion priors, encourages adopting a tripod gait on diverse terrains (see Section \ref{style reward section} for details). Table \ref{tab:reward details} provides the specific reward functions and their scales.
\begin{table}[htbp]
\caption{Reward terms for task tracking, style, and penalty.}
\label{tab:reward details}
\begin{tabular}{@{}ccc@{}}
\toprule
\textbf{Term}      & \textbf{Annotation}      & \textbf{Equation}      \\ \midrule
\multirow{2}{*}{\textbf{Task} $r^g$}    & Linear velocity  & $1*\exp \left(\Vert\mathbf{v}_{t,xy}-\mathbf{v}_{t,xy}^{\rm des}\Vert_2 \big/ 0.15 \right)$        \\
    & Angular velocity & $0.5*\exp \left(\Vert{\omega}_{t,z}-{\omega}_{t,z}^{\rm des}\Vert_2 \big/ 0.15 \right)$              \\ \midrule
\textbf{Style} $r^s$    & D Score    & $1*\max \left[0, 1-0.25\left(d_t^{\rm score}-1\right)^2 \right]$ \\
\midrule
\multirow{10}{*}{\textbf{Penalty} $r^l$}  & Linear velocity  & $-1*{v}_{t,z}^2$ \\
                              & Angular velocity  & $-0.08*\Vert\boldsymbol{\omega}_{t,xy}\Vert_2$ \\ 
                              & Joint torque              & $-2e^{-6}*\Vert\boldsymbol{\tau}\Vert_2$ \\
                              & Joint acceleration        & $-1.5e^{-7}*\Vert\mathbf{\ddot q}\Vert_2$\\
                              & Action rate               & $-0.01*\Vert \mathbf{a}_{t}-\mathbf{a}_{t-1}\Vert_2$ \\ 
                              & Collisions                & $-0.05*n_{collision}$   \\
                              & Joint torque limits       & $-0.05*\Vert\max\left(\left|\boldsymbol{\tau}_t \right|-\boldsymbol{\tau}^{limit},0\right)\Vert_2$      \\
                              & Joint velocity limits     & $-0.5*\Vert\max\left(\left|\boldsymbol{\dot q}_t \right|-\boldsymbol{\dot q}^{limit},0\right)\Vert_2$  \\
                             & Contact force     & $-0.1*\Vert\max\left(\left|\mathbf{f}_t \right|-\mathbf{f}^{limit},0\right)\Vert_2$                   \\ 
\bottomrule
\end{tabular}
\vspace{-0.5em}
\end{table}

We randomize dynamic parameters for both robots and environments to reflect differences between real and simulated conditions. This enhances policy robustness and smooth transfer from simulation to the real world. Details of the parameter randomization are listed in Table \ref{tab:random param}.
\begin{table}[htbp]
\centering
\caption{The range of the randomized parameters.}
\label{tab:random param}
\begin{tabular}{@{}lll@{}}
\toprule
\textbf{Parameters}& \textbf{Range}         & \textbf{Unit}\\ \midrule
Joint Stiffness     & {[}0.8, 1.2{]}$\times$100             &  -          \\
Joint Damping     & {[}0.8, 1.2{]}$\times$2              &  -           \\
Joint Position  & {[}0.6, 1.4{]}$\times$nominal value  &  rad         \\
Link Mass       & {[}0.9, 1.1{]}$\times$nominal value  &  Kg          \\
Payload Mass    & {[}0, 5{]}                           &  Kg          \\
Payload Position& {[}-0.15, 0.15{]} relative to base position &  m       \\
Foot Friction & {[}0.1, 2.5{]}                     &  -          \\
Motor Strength  & {[}0.8, 1.2{]}                       &  -          \\ \bottomrule
\end{tabular}
\vspace{-0.5em}
\end{table}

\subsection{Motion Priors Generation}

The tripod gait is common in hexapod arthropods and is crucial for challenging terrain. To equip our hexapod robot with a high-quality tripod gait, we generate a motion dataset $\mathcal{D}$ on flat ground using TO (see Fig. \ref{fig:method}), which is the most cost-effective way to obtain prior motion data. The resulting trajectories last 8.6 seconds and cover forward, backward, lateral, steering, and combined motions, each maintaining a consistent gait cycle. This ensures the motion data fully corresponds to both the simulated robot and the demonstrator, avoiding extra retargeting \cite{peng2018deepmimic}.
Each state $\boldsymbol{s}_t^{AMP}\in {\mathbb{R}^{61}}$ includes joint positions, joint velocity, base linear and angular velocity, base height relative to the terrain, and foot heights in the base frame. State transitions drawn from $\mathcal{D}$ serve as real samples for discriminator training.

\subsection{Tripod Style Reward Based on Motion Priors}
\label{style reward section}
The style-specific reward promotes a tripod gait similar to the \(\mathcal{D}\) while leaving the robot free to traverse challenging terrain (i.e., it does not force strict imitation). Tripod mode, common in hexapods, ensures the center of gravity remains within the triangular support domain, balancing stability and flexibility. Following \cite{peng2021amp}, we train a discriminator \(D_{\varphi}\) with parameters \(\varphi\) to classify whether each state transition \({T}_s = (\boldsymbol{s}_t, \boldsymbol{s}_{t+1})\) is from the prior dataset or generated by the robot’s policy. If the discriminator detects a difference, it assigns a lower reward, indicating the robot has yet to learn the tripod style. As training progresses, the robot’s transitions become indistinguishable from the prior data, resulting in a higher reward.
The discriminator’s objective is:
\begin{equation} \label{discriminator}
\begin{split}
\mathop{\arg\min} \limits_{\varphi} &\mathbb{E}_{ {T}_s \sim \mathcal{D}} \left[\left(D_{\varphi}  ({T}_s)  - 1\right)^2\right] + \mathbb{E}_{  {T}_s \sim \pi} \left[\left(D_{\varphi}  ({T}_s) + 1\right)^2\right] \\ &+ \frac{{{\alpha ^{gp}}}}{2}{_{{T}_s\sim{\cal D}}}\left[ \left \Vert {{\nabla _\varphi }{D_\varphi } ({T}_s)} \right\Vert_2 \right],
\end{split}
\end{equation}
where the first two terms use a least square GAN formulation to minimize the Pearson divergence between transitions from $\pi$ and \(\mathcal{D}\). To stabilize training, a gradient penalty is introduced in the second term \cite{peng2021amp}, controlled by \(\alpha^{gp}\). The tripod style reward is then defined as:
\begin{equation} \label{style reward}
r_t^s\left[{T}_s\sim \pi\right] = \max \left[0, 1-0.25\left({D_\varphi } ({T}_s)- 1\right)^2 \right],
\end{equation}
and is scaled to the range $\left[0, 1\right]$.

\section{NETWORK DESIGN AND TRAINING}

\subsection{Network Architecture}

We establish an asymmetric Actor-Critic RL framework: the Critic network receives privileged data and terrain details via two encoders to evaluate the current policy’s actions, while the Actor network relies solely on observable measurements (velocity commands, previous actions, and proprioceptive observations) for deployment. We encode terrain information $\boldsymbol{i}_t^e$ into a 16-dimensional latent variable $\boldsymbol{l}_t^e$ using a terrain encoder \(g_e\), and encode privileged data $\boldsymbol{s}_t^p$ into an 8-dimensional latent variable $\boldsymbol{l}_t^p$ using a privileged encoder \(g_p\). A three-layer Critic MLP then processes these latent representations and the observable data to produce target values $V_t$ for advantage estimation.

Because it is difficult to obtain accurate linear velocity on real robots, we introduced a state estimator within the Actor network that computes linear velocity from the last five proprioceptive observations $\boldsymbol{o}_{t-N+1}^p,...,\boldsymbol{o}_{t-1}^p,\boldsymbol{o}_t^p,(N=5)$. We also designed a short-term memory encoder to compress these past observations into a latent variable $\boldsymbol{h}_t$, allowing the robot to infer terrain characteristics from its history. The observable variables, estimated velocity, and the latent representation of past states are then passed to a low-level MLP, which produces the policy action $\boldsymbol{a}_{t}$.
The discriminator \(D_{\varphi}\) is a simpler network with two hidden layers and a linear output. Further details can be found in Table \ref{tab:net_arc}.
\begin{table}[htbp]
\renewcommand{\arraystretch}{1.25}
\setlength{\tabcolsep}{3pt}
\centering
\caption{Network architecture for RL training framework.}
\label{tab:net_arc}
\begin{tabular}{@{}lllll@{}}
\toprule
\textbf{Module} & \textbf{Inputs} & \textbf{Hidden Layers} & \textbf{Outputs} \\ \midrule
Estimator (MLP) & ${O}_{t-4}^{p},...,{O}_{t}^{p}$   & {[}64, 32{]}     &$\hat{v}_t$  \\
Memory (MLP)   & ${O}_{t-4}^{p},...,{O}_{t}^{p}$& {[}512, 256, 128{]}      & $h_t$ \\
Low-Level (MLP) & $cmd,a_{t-1},o_t^p,\hat{v}_t,h_t$   & {[}256, 128, 64{]}     &$a_t$  \\
$g_p$ (MLP)  & $s_t^p$    & {[}64, 32{]}            & $l_t^p$  \\
$g_e$ (MLP)  & $i_t^e$   & {[}256, 128{]}          & $l_t^e$ \\
Critic (MLP)    & $cmd,a_{t-1},o_t^p,l_t^p,l_t^e$             & {[}512, 256, 128{]}      & $V_t$  \\
$D_{\varphi}$ (MLP)& $s_t^{AMP},s_{t+1}^{AMP}$ & {[}1024, 512{]}& $d_t^{\rm score}$ \\ \bottomrule
\end{tabular}
\end{table}

\subsection{Training}

We train the policy using Proximal Policy Optimization (PPO) \cite{schulman2017proximal} with privileged state and terrain data. At the start of each episode, the robot receives random velocity commands $\boldsymbol{v}_t^{\rm des}$,
representing longitudinal, lateral, and yaw velocities. Following the terrain curriculum~\cite{wjz}, the yaw velocity is provided directly for efficient tracking. The policy network estimates the robot's linear velocity $\hat{\boldsymbol{v}}_t$ through supervised learning using privileged information (see Fig. \ref{fig:method}).
We update the discriminator and policy networks concurrently. Specifically, we randomly extract state transition pairs $\boldsymbol{T}_{s}^{p}=\left(\boldsymbol{s}_t^{p},\boldsymbol{s}_{t+1}^{p}\right)$ from prior data, while the policy generates its own pairs $\boldsymbol{T}_{s}^{\pi}=\left(\boldsymbol{s}_t^{\pi},\boldsymbol{s}_{t+1}^{\pi}\right)$.
The discriminator $D_{\varphi}$ evaluates these pairs and outputs ${D_\varphi } ({T}_s)$, which is used to compute the $r^s_t$. The policy learns the prior motion style by generating actions that deceive the discriminator, which is updated simultaneously to better distinguish between the prior data and the agent's behavior.

\section{SIMULATIONS AND EXPERIMENTS}

\textbf{Simulation:} We created the terrains in the IsaacGym and trained 4096 robots simultaneously \cite{rudin2022learning}. Each episode involved 1000 steps over 20 seconds, with early termination if the condition was met. The policy ran at a control frequency of \SI{50}{\hertz}. We conducted 50,000 episodes, and the training took about 35 hours on a NVIDIA RTX 3090Ti GPU.

\textbf{Hardware:} Our hexapod robot has a symmetrical design with six legs: right front (RF), right middle (RM), right rear (RR), left front (LF), left middle (LM), and left rear (LR). Each leg has three degrees of freedom, including the hip, thigh, and shank joints. To prevent leg collisions and increase the support area, the middle legs are extended 13.7 cm outward compared to the front and rear legs. The robot weighs 25.5 kg and stands 30 cm tall.

\subsection{Ablation Study for the Design of the Reward Terms}

To ascertain the necessity of each type of reward term, we trained three policies considering different combinations of the rewards, including $r^g_t+r^s_t$, $r^g_t+r^l_t$, and $r^g_t+r^s_t+r^l_t$.

\begin{figure}[tbp]
\includegraphics[width=\linewidth]{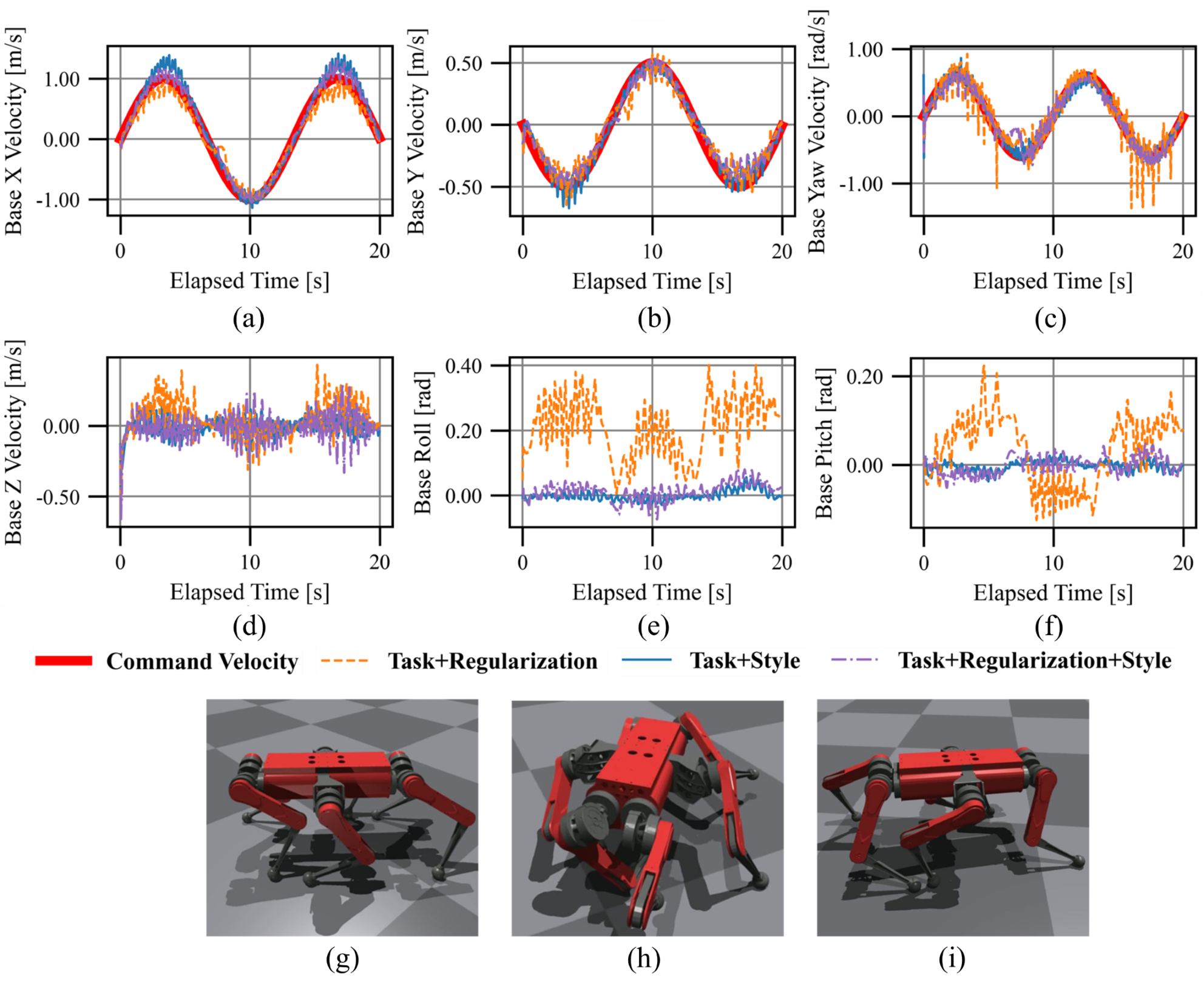}
\centering
\caption{Comparison of three policies in terms of ability to track sinusoidal velocity commands in the simulation. (a)-(c) Base velocity tracking in x, y, yaw directions. (d)-(f) Base velocity deviations in z-axis, and orientation deviations along the x, y axes. (g) Locomotion guided by $r^g_t+r^s_t$. (h) Locomotion guided by $r^g_t+r^l_t$. (i) Locomotion guided by $r^g_t+r^s_t+r^l_t$.}
\label{fig:sin track}
\vspace{-8pt}
\end{figure}

We first analyzed the locomotion behavior of the three policies on flat terrain. Fig. \ref{fig:sin track} compares the tracking performance of the policies on flat ground with velocity commands in simulation. Fig. \ref{fig:sin track}(a), (b), and (c) show the performance of the policies on sinusoidal velocity commands in the x and y directions and yaw angular velocity. The results show that the policy guided by $r^g_t + r^l_t$ exhibits significant jitter and deviation in velocity tracking, leading to unnatural behavior, as seen in Fig. \ref{fig:sin track}(h).
Fig. \ref{fig:sin track}(d), (e), and (f) compare the stability of the policies in the z-direction linear velocity and roll and pitch angles. The severe deviation of the curve guided by $r^g_t + r^l_t$ shows that without the style reward, the policy fails to suppress movement in unexpected directions. This suggests that the style reward $r^s_t$ helps the policy learn behaviors that better capture the reference tripod gait.

\begin{figure}[!htb]
\vspace{-0.5em}
  \centering
  \subfigure[]{\includegraphics[width = 0.48\hsize]{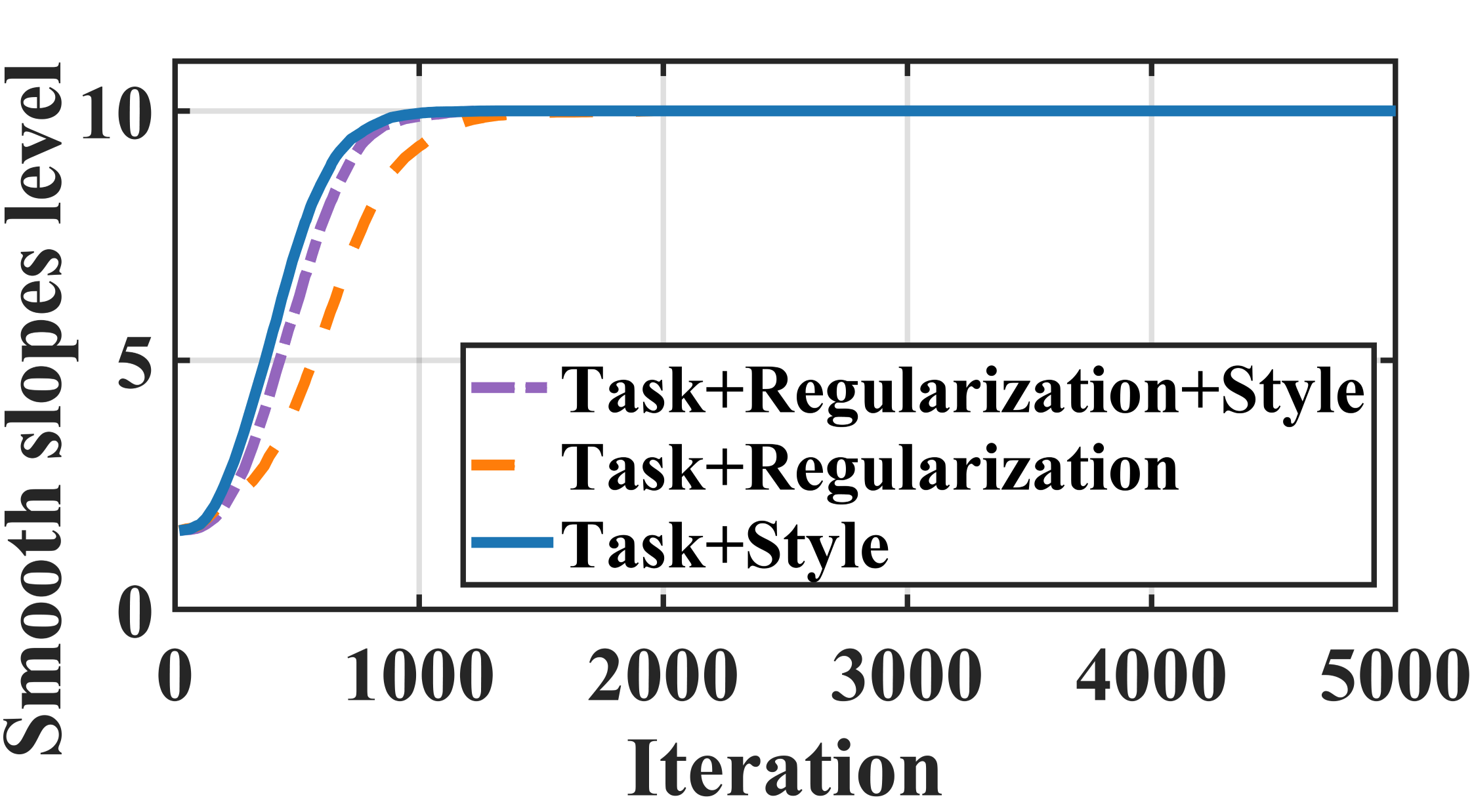}\label{subfig:Smoothslopelevel}}
  \subfigure[]{\includegraphics[width = 0.48\hsize]{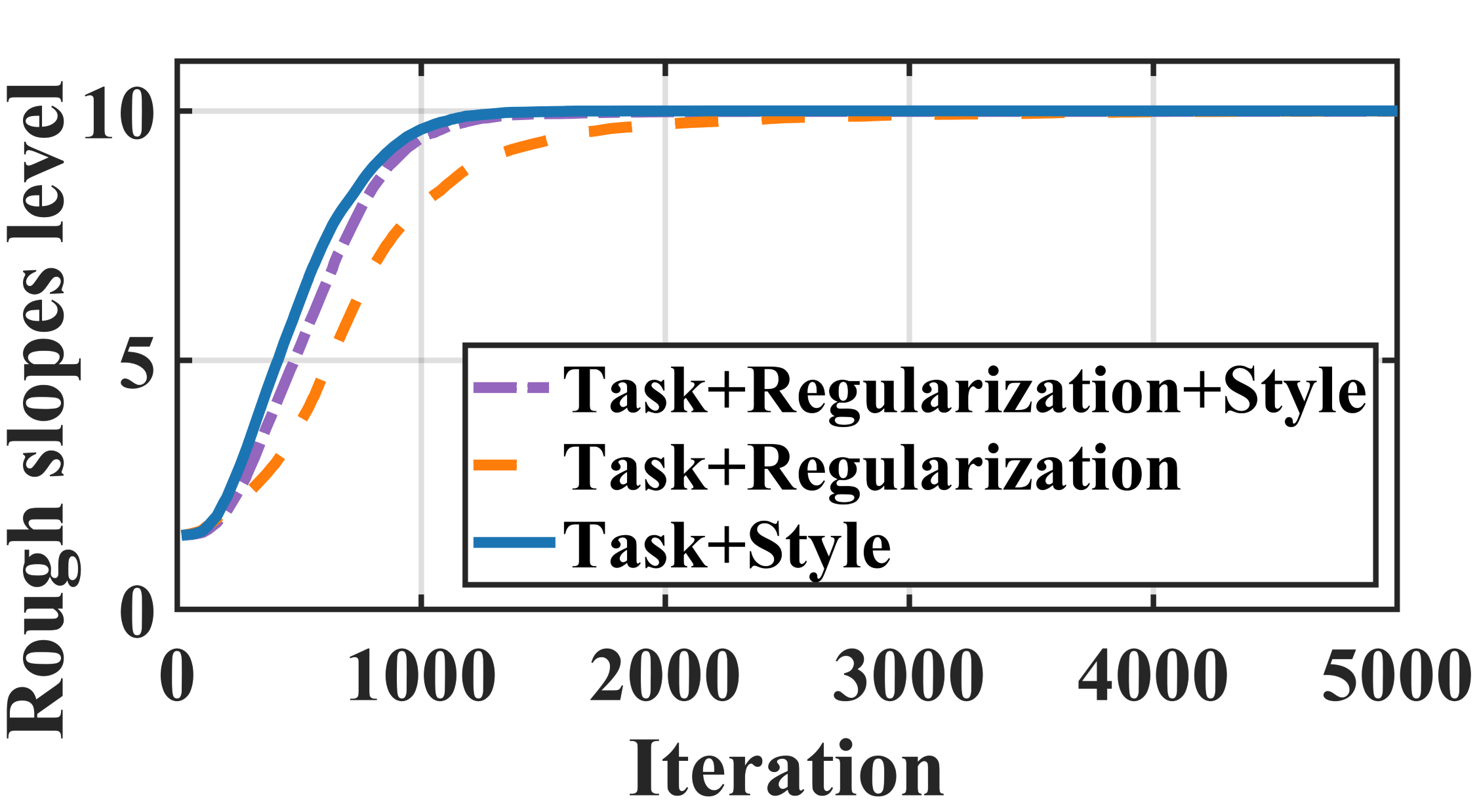}\label{subfig:Roughslopelevel}}
  \subfigure[]{\includegraphics[width = 0.48\hsize]{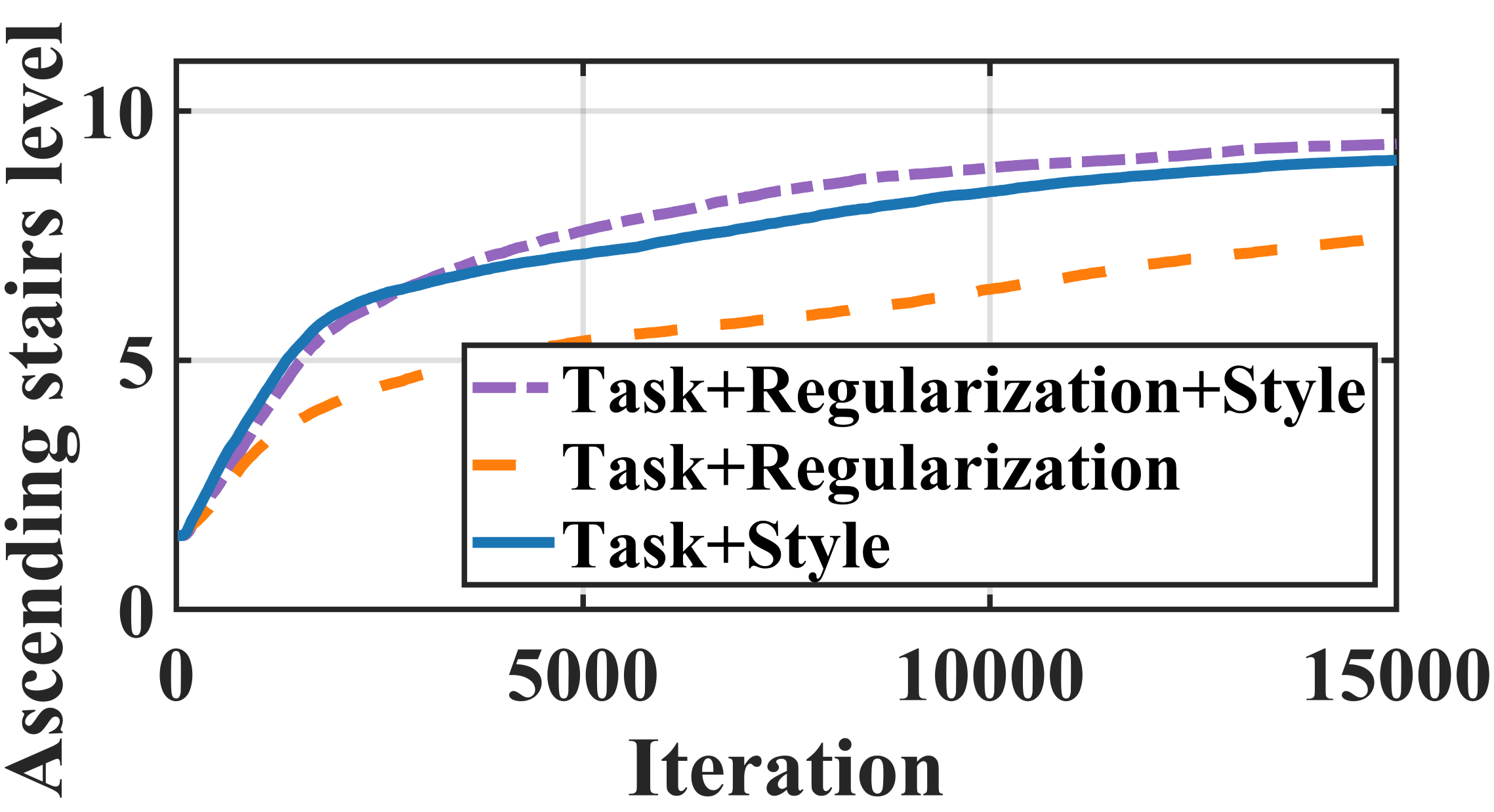}\label{subfig:Ascendingstairslevel}}
  \subfigure[]{\includegraphics[width = 0.48\hsize]{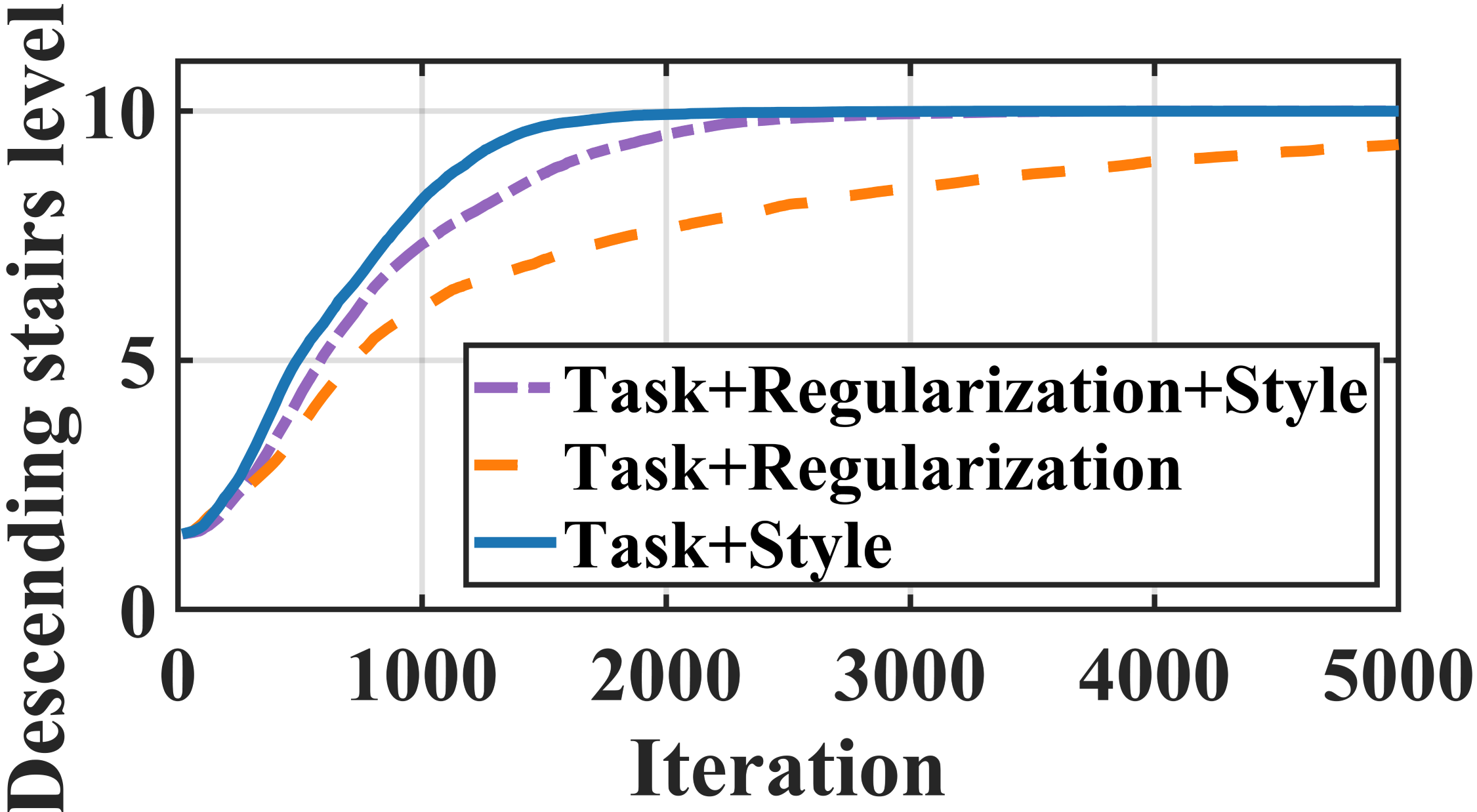}\label{subfig:Descendingstairslevel}}
  \caption{Comparison of policies in terms of ability to travel different terrains.}
\label{fig:terrains trav}
\vspace{-0.5em}
\end{figure}

Next, we compared the traversability of the three policies across various terrains. In Fig. \ref{fig:terrains trav}, the vertical axis shows the terrain difficulty, and the horizontal axis represents iterations. The results show that the policies guided by $r^g_t + r^s_t + r^l_t$ and $r^g_t + r^s_t$ enable the robot to navigate more difficult terrains faster and reach higher levels. Specifically, for challenging terrains like stairs, the policies with $r^g_t + r^s_t + r^l_t$ and $r^g_t + r^s_t$ perform better than the $r^g_t + r^l_t$ policy, as seen in Fig. \ref{subfig:Ascendingstairslevel} and \ref{subfig:Descendingstairslevel}. This suggests that relying solely on task rewards and penalties may lead to abnormal behavior, limiting traversal of complex terrain. The style reward helps the robot learn more natural behaviors and explore its motion capabilities. Additionally, the policy with $r^g_t + r^s_t$ performs better in the early stages, but $r^g_t + r^s_t + r^l_t$ helps the robot navigate more difficult terrain in later stages, as shown in Fig. \ref{subfig:Ascendingstairslevel}.

\subsection{Robustness Experiments}

We also trained and compared other advanced RL-based controllers and OCS2-based MPC controllers. We assessed each controller's performance on flat terrain and its ability to navigate challenging terrains. These controllers are all blind locomotion controllers that rely on proprioception, including:
\begin{enumerate}
\item \textit{Baseline \cite{rudin2022learning}:} A policy trained without privileged access to information about the environment.
\item \textit{Concurrent \cite{ji2022concurrent}:} A policy is trained without terrain information as input for the actor network, and concurrently trained with an estimator network that estimates the body state.
\item \textit{RMA \cite{kumar2021rma}:} A policy trained using a teacher-student framework without any expert priors.
\item \textit{MPC \cite{FARSHIDIAN7989016}:} An MPC controller based on OCS2 fine-tunes the leg lift height, body height, and gait.
\end{enumerate}

We used 5 random seeds and the same low-level network.

To test robustness, we applied random disturbances to the robot on flat ground in simulation. Specifically, we applied velocity perturbations along the three coordinate axes, ranging from small to large magnitudes. These perturbations affected the robot's center of mass position at one-second intervals until a termination condition was met. Table \ref{tab:velocity_disturbances} shows the maximum velocity disturbances each controller could handle without causing the robot to fall. Results show the robot is least disturbance‑tolerant along the Y‑axis; the table below gives its Y‑axis tolerance range.
\begin{table}[htbp]
        \renewcommand{\arraystretch}{1.3}
        \caption{The controllers' velocity disturbance tolerance range.}
        \centering
        \label{tab:velocity_disturbances}
            \begin{tabular}{@{}cc@{}}
                \toprule
                \textbf{Controllers} & \textbf{Disturbances [Min, Max] (m/s)}  \\ \midrule
Ours                  & {[}\textbf{-0.803}, \textbf{0.803}{]}       \\
RMA                   & {[}-0.738, 0.738{]}                         \\
Concurrent            & {[}-0.463, 0.463{]}                         \\
Baseline              & {[}-0.201, 0.201{]}                         \\
MPC                   & {[}-0.112, 0.112{]}                         \\
\bottomrule
\end{tabular}
\end{table}

The results showed controllers could regain stability after disturbances. Exceeding the threshold caused a loss of control, highlighting the different robustness among the controllers. Notably, our method demonstrated superior robustness, handling larger disturbances better than the others.

\subsection{Indoor and Outdoor Experiments}

As shown in Fig. \ref{fig:locomotion}, we tested the robot on stairs ranging from 3 cm to 20 cm in height and on slopes with gradients from 5° to 30°. The robot moved at 0.3 m/s for 10 s. Success was defined as completing the tasks—ascending/descending stairs or traversing slopes—without falling. We conducted 10 tests for each controller and calculated the success rate.

\begin{figure}[!htb]
  \centering
  \subfigure[]{\includegraphics[width = 0.48\hsize]{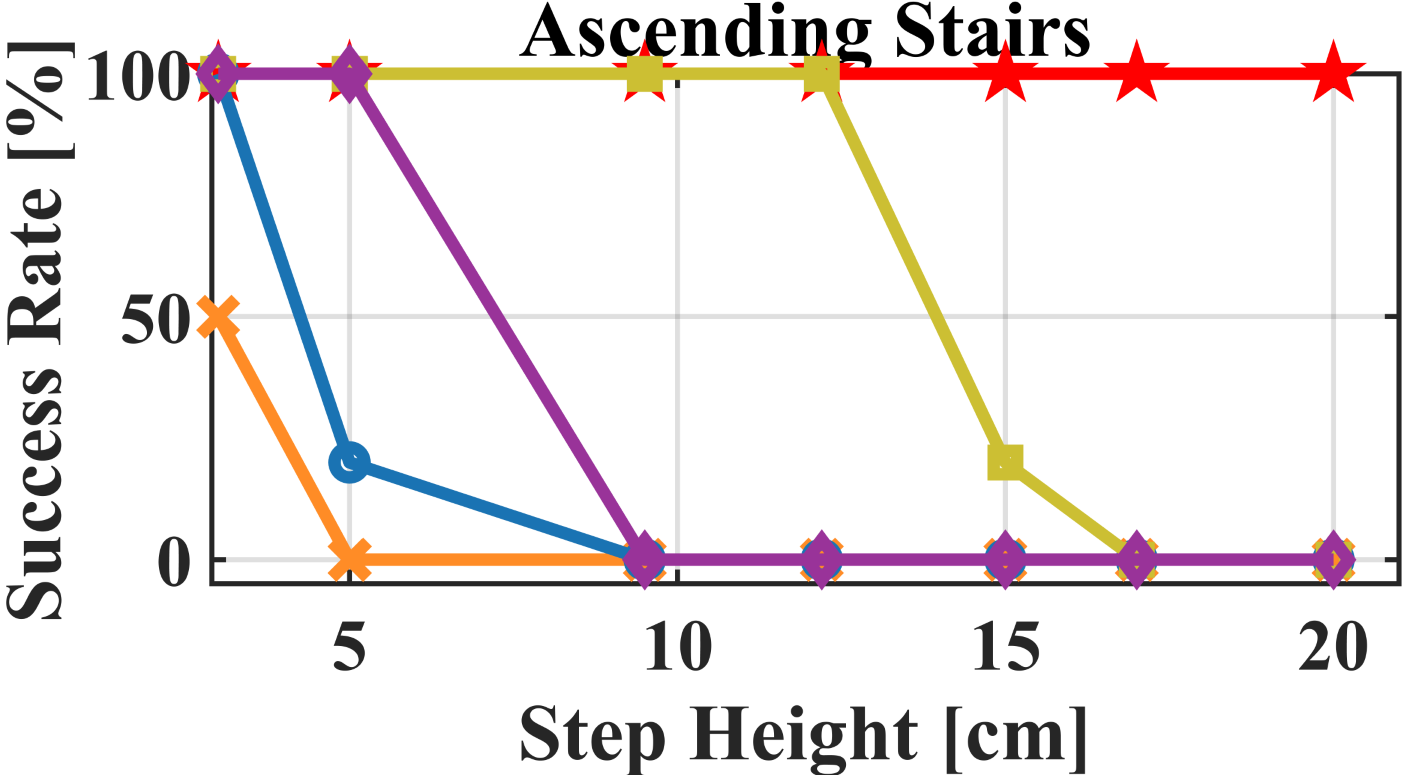}\label{subfig:ascending}}
  \subfigure[]{\includegraphics[width = 0.48\hsize]{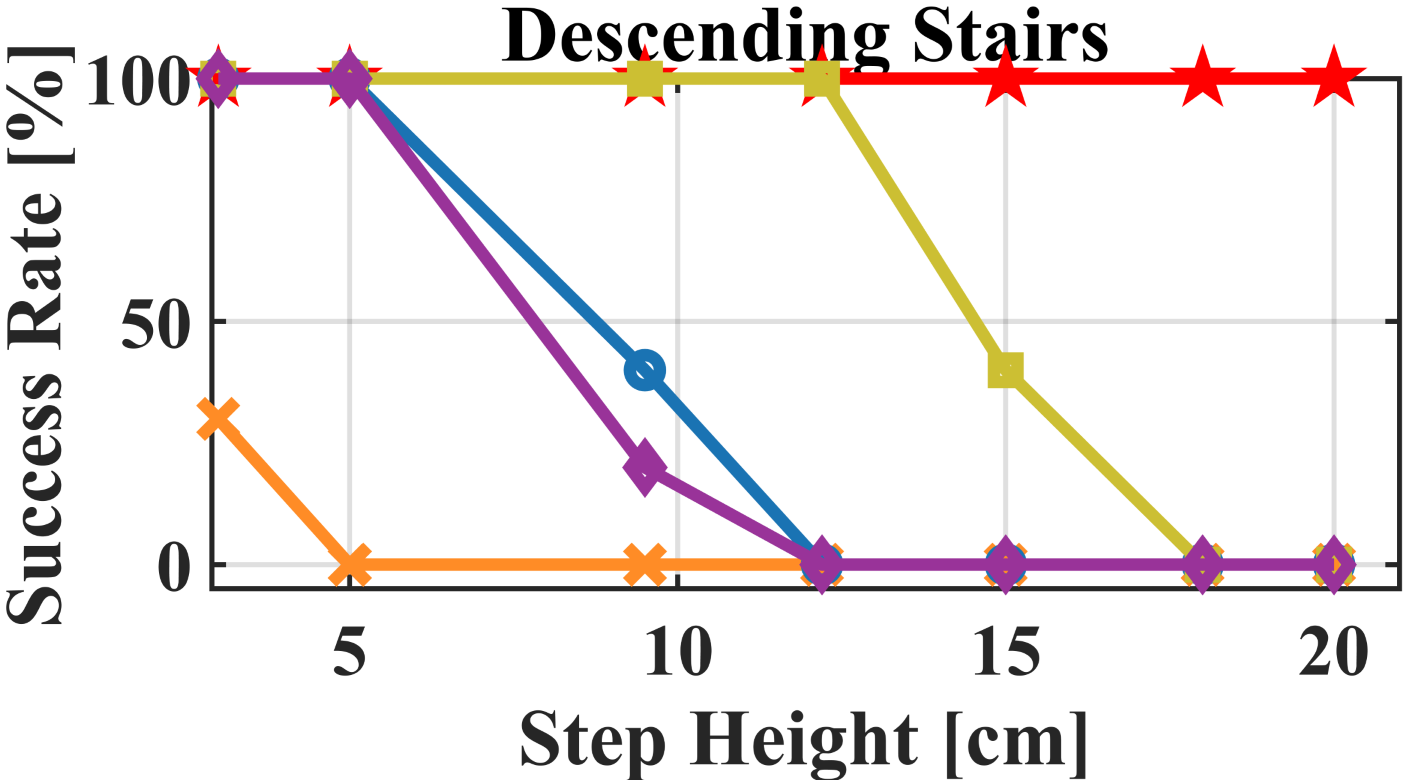}\label{subfig:descending}}
  \subfigure[]{\includegraphics[width = 0.48\hsize]{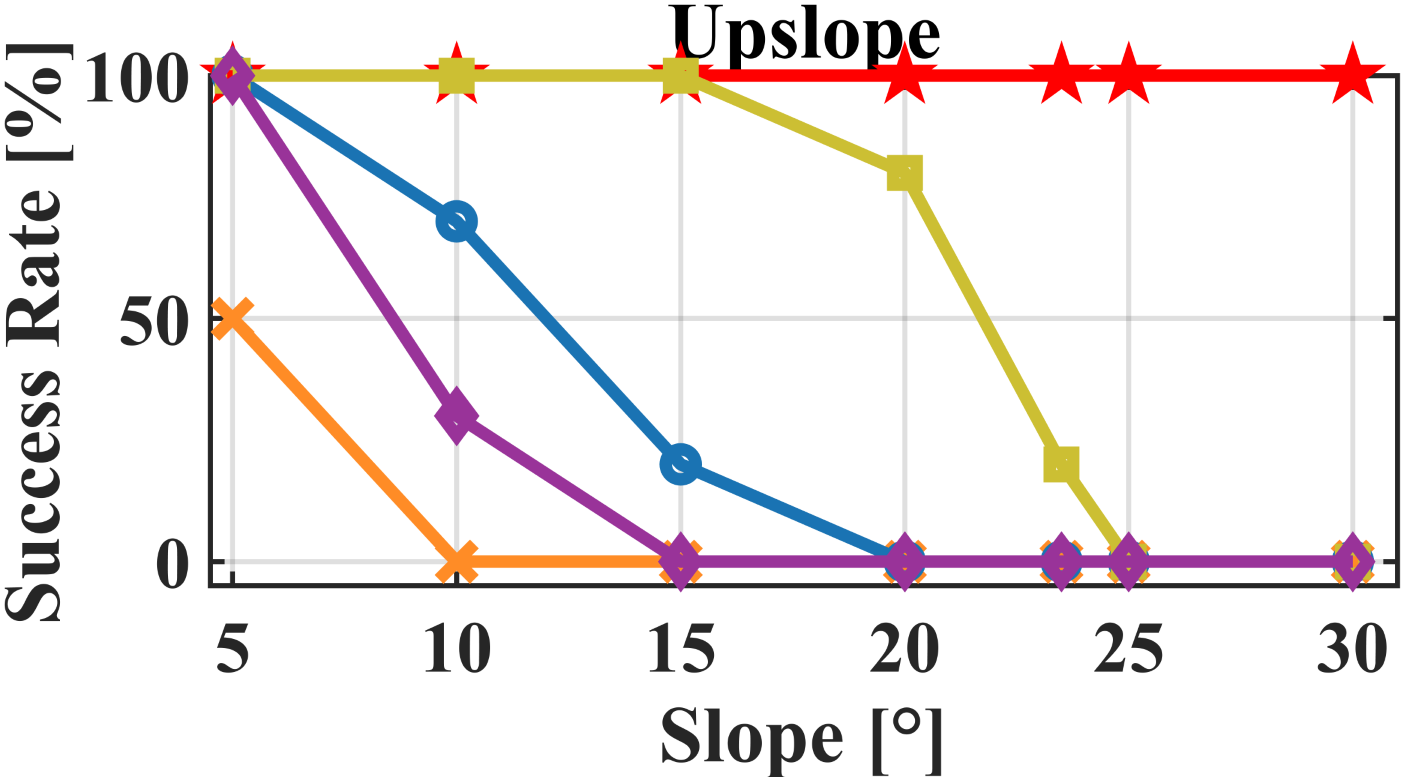}\label{subfig:slope}}
  \subfigure[]{\includegraphics[width = 0.48\hsize]{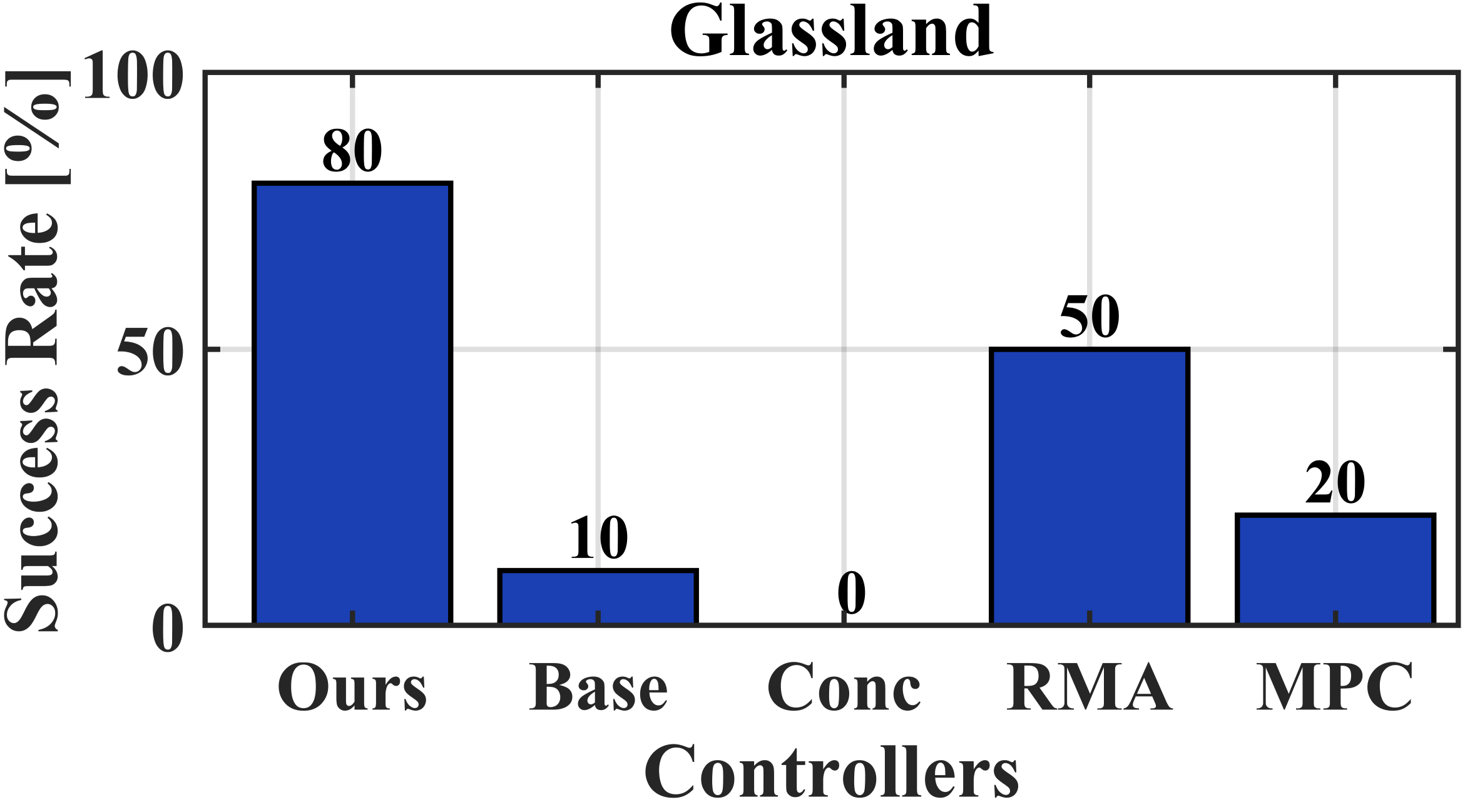}\label{subfig:glassland}}
  \subfigure{\includegraphics[width=2.2in]{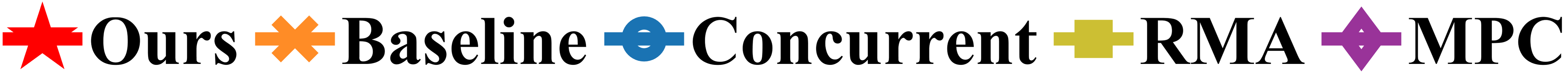}}
  \caption{Success rates of different controllers in different terrains}
\label{fig:success_rate}
\vspace{-0.5em}
\end{figure}

As shown in Fig. \ref{fig:success_rate}(a)-(c), our controller successfully navigated all terrains. RMA can access terrain information during teacher policy training, allowing some adaptation to terrains. However, its fixed low-level network updates limit adaptability to more complex terrains. The asymmetric Actor-Critic method addresses this by continuously updating the low-level network. Additionally, RL controllers trained with Baseline, Concurrent, or the MPC controller struggled to adapt to complex terrains without terrain information.

In the outdoor test, we navigated the robot at 0.5 m/s across a flower bed with a 15 cm step and over approximately 46 m of uneven grassland, as shown in Fig. \ref{fig:locomotion}. Success was defined as crossing the flower bed without falling. We conducted 10 tests for each controller and calculated the success rate. As shown in Fig. \ref{subfig:glassland}, our controller consistently outperformed the others. This demonstrates its ability to adapt to soft, uneven grass terrain, not encountered in simulation, with the memory encoding network’s terrain inference helping the controller adjust to complex terrain.

\section{CONCLUSIONS}

In this paper, we propose a novel approach that combines motion priors with reinforcement learning (RL) algorithms. An RL controller is trained with an adversarial discriminator using these motion priors. This method enables the hexapod robot to perform natural and robust blind locomotion in complex terrains. Simulations and experiments show that the learned policy transfers successfully to the real robot, demonstrating natural gaits and strong robustness without visual input in challenging environments.











\bibliographystyle{./IEEEtran} 
\bibliography{./main}


\end{document}